\documentclass[letterpaper, 10 pt, conference]{ieeeconf}  

\IEEEoverridecommandlockouts                              

\usepackage[utf8]{inputenc}
\usepackage{graphicx}
\usepackage{amsmath,amsfonts,booktabs,cite} 
\usepackage{gensymb} 
\usepackage{siunitx,textcomp} 
\usepackage[hidelinks]{hyperref} 
\usepackage{cleveref} 
\usepackage{tabularx}
\usepackage{pbox}
\let\labelindent\relax
\usepackage[inline]{enumitem} 
\usepackage[nolist]{acronym} 
\usepackage{multirow}
\usepackage{url}
\usepackage{algorithm}
\usepackage[noend]{algpseudocode}
\usepackage{subfigure}
\usepackage{flushend}

\usepackage{caption}
\def\figref#1{Fig.~\ref{#1}}

\def\eqref#1{Eq.~(\ref{#1})}

\newcommand\etal{~\emph{et al. }}

\newsavebox{\twosubbox}

\crefname{algocf}{alg.}{algs.}
\Crefname{algocf}{Algorithm}{Algorithms}

\title{\LARGE \bf
Spatiotemporal Attention Enhances Lidar-Based Robot Navigation in Dynamic Environments
}

\author{Jorge de Heuvel \and Xiangyu Zeng \and Weixian Shi \and Tharun Sethuraman \and Maren Bennewitz
\thanks{
	All authors are with the Humanoid Robots Lab, University of Bonn, Germany.
	Maren Bennewitz and Jorge de Heuvel are additionally with the Lamarr Institute for Machine Learning and Artificial Intelligence, Germany. 
	This work has partially been funded by the Deutsche Forschungsgemeinschaft (DFG, German Research Foundation) under the grant number BE~4420/2-2~(FOR 2535 Anticipating Human Behavior).
	}
}
\begin{document}
	
\maketitle
\thispagestyle{empty} 
\pagestyle{empty}

\begin{abstract} 
Foresighted robot navigation in dynamic indoor environments with cost-efficient hardware necessitates the use of a lightweight yet dependable controller. 
So inferring the scene dynamics from sensor readings without explicit object tracking is a pivotal aspect of foresighted navigation among pedestrians.
In this paper, we introduce a spatiotemporal attention pipeline for enhanced navigation based on 2D~lidar sensor readings. 
This pipeline is complemented by a novel lidar-state representation that emphasizes dynamic obstacles over static ones.
Subsequently, the attention mechanism enables selective scene perception across both space and time, resulting in improved overall navigation performance within dynamic scenarios.
We thoroughly evaluated the approach in different scenarios and simulators, finding excellent generalization to unseen environments. 
The results demonstrate outstanding performance compared to state-of-the-art methods, thereby enabling the seamless deployment of the learned controller on a real robot.
\end{abstract}

\vspace*{-0.5em}
\section{Introduction}
\vspace*{-0.5em}
\label{sec:intro}
The deployment of mobile service robots around our living areas to improve humans’ daily life quality, such as by performing house chores, or carrying out delivery tasks, is an ongoing evolution.
For seamless navigation among humans, learning-based navigation controllers represent the forefront of research.
A key performance requirement is usually an information-dense representation of the dynamic scene, e.g., with explicitly tracked pedestrians \cite{katyal2020intent}.
However, when transitioning away from test and training simulations to the real robot, complex fusion from multiple sensors and hardware-heavy post-processing steps are required to achieve such information-dense dynamics representations\cite{wang2022group, wang2023multi, wang2021pointaugmenting}.
Here, also feature-rich but costly 3D~lidar sensors are appealing \cite{li2021openstreetmap, himmelsbach2010fast}.
On the other side of the spectrum, many studies focus on learning-based navigation among dynamic obstacles of known position to avoid sensor-based pedestrian tracking \cite{chen2019crowd, chen2020relational}. 
These approaches suffer from a reality gap that hinders generalization to the real world\cite{salvato_crossing_2021, chen_understanding_2022}.
Following the demand for improved reactive local planners, as recently emphasized by Xiao \etal \cite{xiao_motion_2022}, the need for sensor-based lightweight but reliable perception and navigation pipelines emerges that rendundantizes explicit obstacle tracking.

A possible solution is the use of 2D~lidar sensors that provide accurate obstacle information within the moving plane of mobile robots \cite{tai2017virtual}.
They operate independently of lighting conditions, enabling both day and night operation.
But without data such as colors or contours, explicitly tracking objects instances like pedestrians only by their leg profiles from 2D~lidar readings is a hard task \cite{qin2016spatial, song20192d}.
Furthermore, the robot's self-movement makes static objects appear dynamic between lidar scans.

\begin{figure}[!t]
	\centering
	\includegraphics[width=1.0\linewidth]{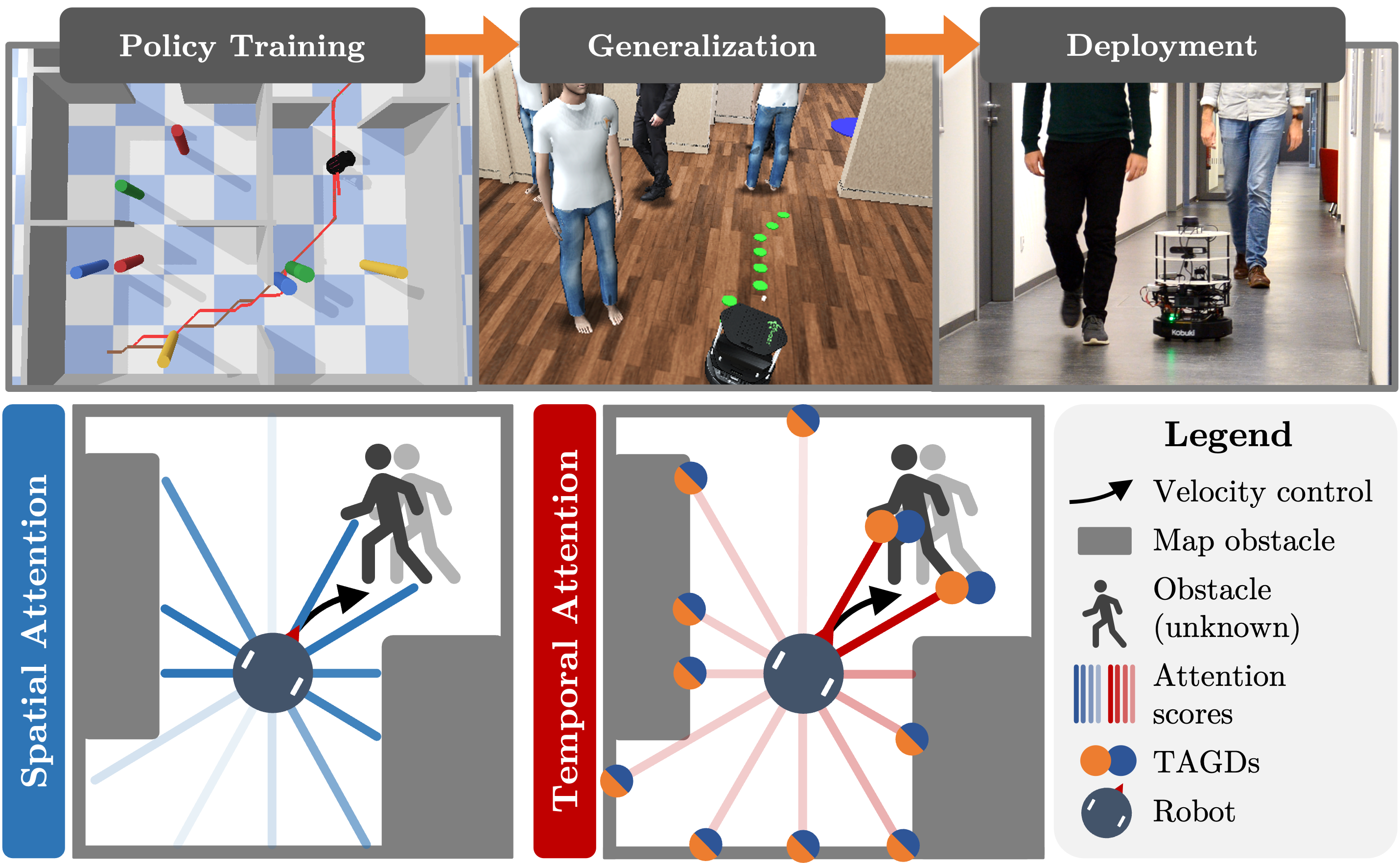}
	\caption{
		Our pipeline for learning a robot navigation controller based on lidar.
		Two attention mechanisms reason about the importance of individual lidar sectors with respect to known and unknown dynamic obstacles.
		Our Temporal Accumulation Group Descriptors (TAGD) reveal moving obstacles from subsequent lidar scans affected by robot self-motion.
		}
	\label{fig:motivation}
\end{figure}

While most current methods leverage convolutional neural networks (CNNs) to process and extract features from lidar data \cite{perez-darpino_robot_2021, liang_realtime_2020}, a recent appealing idea to tackle these sensor-implicit obstacle representations is selective attention on a collision-relevant subsectors of the lidar data \cite{de_heuvel_subgoal-driven_2023}.
Especially when a temporal observation sequence provides dynamic scene information, selective attention on moving obstacles can be beneficial.

To address this, we introduce a novel feature extraction technique tailored for 2D~point clouds, incorporating both spatial and temporal attention across the sensor readings. This approach distills critical navigational information, offering a more robust solution for learning-based navigation in dynamic indoor environments.
We demonstrate better than state-of-the-art generalization to unseen navigation scenarios and enable a smooth sim-to-real transfer of the learned policy, as we will be able to demonstrate in the experiments.

In summary, the main contributions of our work are:

\begin{itemize}
	\item A deep reinforcement learning-based (DRL) navigation controller that learns dynamic obstacle avoidance implicitly from 2D~lidar readings only. 
	\item A spatiotemporal attention module that infers the relative importance of different observation sectors with respect to proximity and obstacle motion trends.
	\item A novel 2D~lidar observation representation highlighting dynamic obstacles over the robot's self-motion called temporal accumulation group descriptor (TAGD).
\end{itemize}

\vspace*{-0.5em}
\section{Related Work}
\vspace*{-0.25em}

Where mobile robot navigation decomposes into global path planning and local obstacle avoidance, the latter can be tackled with traditional and learning-based approaches.
While traditional approaches such as the popular dynamic window approach (DWA) \cite{fox1997dynamic, brock1999high} have been advanced with motion prediction \cite{missura_predictive_2019}, they come with the difficulties to avoid C-shaped or dynamic obstacles, or the necessity for re-tuning in different environments \cite{xiao2020appld}. 

\label{sec:related}

\vspace*{-0.25em}
\subsection{Learning-based navigation}\label{ssec:ss2}
\vspace*{-0.25em}
\begin{figure*}[!ht]
	\centering
	\includegraphics[width=0.95\linewidth]{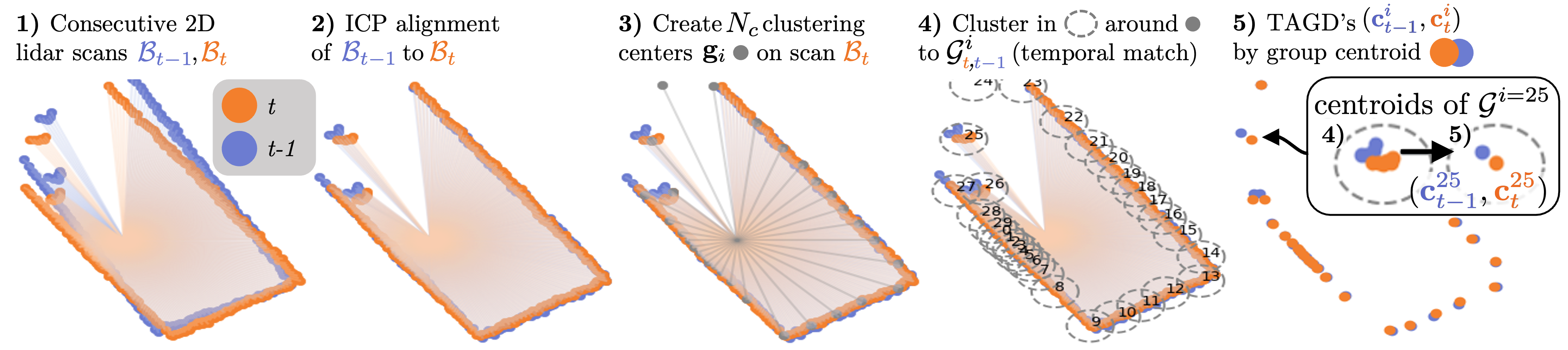}
	\caption{
		Schematic of the TAGD generation process.
		The ICP alignment of two subsequent lidar scans (1) in 2D~Cartesian coordinates reduces the effect of robot self-movement (2).
		This allows better differentiation between dynamic obstacles and static obstacles.
		The aligned scan is grouped and clustered around ray-cast centers (3).
		From the clustered points (4), the position difference of the centroid from both time steps reveals a moving obstacle (5).
	} 
	\label{fig:tag_descriptor}
\end{figure*}
Deep learning-based methods \cite{shabbir2018survey} appeal with decent generalization performance and less tedious fine-tuning as compared to hard-coded controllers.

Especially reinforcement learning-based (RL) methods have successfully been applied to motion planning \cite{tai2017virtual, chiang2019learning, de_heuvel_learning_2023-1, de_heuvel_subgoal-driven_2023, faust2018prm}.
These works however do not embed dynamic scene understanding, thus limiting the agent's capability around walking pedestrians. 
Methods like SARL \cite{chen2019crowd} or MP-RGL \cite{chen2020relational} capture interactions between robot and humans with excellent results, but rely on the known velocity of humans. 
Others infer or forecast human behavior by predicting their long-term goals, or by predicting their future motion and activities \cite{bayoumi2016learning}, often by employing 3D~lidar or RGB(D) cameras \cite{pfrunder2017real, fan2019getting, liang_realtime_2020}.
The challenge arises with our aim to learn time-series motion trends for scene-dynamics aware navigation from 2D~lidar readings in an end-to-end manner.

\vspace*{-0.5em}
\subsection{Point Cloud Feature Extraction}\label{ssec:ss3}
\vspace*{-0.25em}

For the feature extraction in a deep RL navigation task, the spatial nature of lidar point cloud data suggests convolutional neural networks (CNN) as natural fit \cite{perez-darpino_robot_2021, liang_realtime_2020}.
Here, reducing the input dimensionality into a sparse encoding is a pivotal step.
Taking into account the temporal dimension for scene dynamics understanding, individual lidar scans may be CNN-processed separately, followed by a  multilayer perceptron (MLP) for joint extraction \cite{liang_realtime_2020}.
With PointNet \cite{charles_pointnet_2017}, a high performing network architecture for 3D point cloud registration has been proposed that was recently put to test in a short-horizon RL-based robotic manipulation task \cite{ling_efficacy_2023}.
For obstacle pose and dynamics estimation, using a point cloud segmentation approach represents a viable avenue \cite{himmelsbach2010fast}.
Looking at the non-learning-based domain, obstacle tracking from point cloud data has been presented before \cite{kraemer_lidar-based_2018, chen_real-time_2021}.
With the advent of transformer models, the self-attention operator's invariance to cardinality and permutation of input data has proven to be a useful property for point cloud inference \cite{zhao_point_2021-1}.
Building on this foundation, our work leverages attention-based 2D~lidar feature extraction.
This approach enhances deep RL-based local-obstacle avoidance, while integrating high-level guidance from a conventional path planner.

\vspace*{-0.5em}
\section{Problem Statement and Assumptions}
\vspace*{-0.25em}
In this work, we consider a differential-wheeled robot pursuing a global goal in a cluttered and dynamic indoor environment, compare Fig.~\ref{fig:network_architecture}a).
A map of the empty environment is available for global path planning via A*.
Static or dynamic pedestrians however are unknown obstacles to the robot. 
Also, the pedestrians at different speeds move rigorously without avoiding the robot in their motion, in contrast to other social navigation studies \cite{perez-darpino_robot_2021}.
Therefore, smart and foresighted local collision avoidance is entirely up to the robot.
The controlling agent has access to subsequent 2D~lidar readings and upcoming path waypoints as observations, which it maps to linear and angular velocity commands.
We formulate the task as in a learning-based manner and apply off-policy DRL.
In summary, the proposed controller should be able to achieve two tasks: 
1) Pursue the global goal through guidance of the computed path and 2) effectively avoid dynamic obstacles on a local scale.

\vspace*{-0.5em}
\section{Our Approach}
\vspace*{-0.25em}
\label{sec:approach}
This section explains our novel temporal accumulation group descriptor for lidar readings and subsequently the learning framework.

\vspace*{-0.5em}
\subsection{Temporal Accumulation Group Descriptor (TAGD)}
\vspace*{-0.25em}
It is inherently difficult to capture motion trends of moving obstacles from consecutive 2D~lidar readings when the robot is in motion. 
To reveal moving obstacles over static ones without explicit obstacle tracking, we introduce our novel TAGD.
We assume lidar scans to be recorded at a constant frequency of $1/\Delta t$ with a range of $d_\text{max}$.
Our approach is described in Algorithm \ref{algo:tagd} with a visualization of all major steps is shown in Fig.~\ref{fig:tag_descriptor}.
We start with the min-pooled 2D~lidar points $\mathcal{B}_{t-1}, \mathcal{B}_{t}$ with $N$ points each.
To eliminate the impact of robot rotation and translation, ICP~\cite{arun1987least} aligns $\mathcal{B}_{t-1}$ to $\mathcal{B}_{t}$ in the transformed point set $\mathcal{B}^{'}_{t-1}$~(Fig.~\ref{fig:tag_descriptor}.1).
For static obstacles, the points now match up while their positions misalign for dynamic obstacles (Fig.~\ref{fig:tag_descriptor}.2).
For spatial clustering and subsequent temporal matching, clustering group centers $g^i$ are formed along $N_c$ uniformly cast rays by determining the robot-closest point within an angular threshold $\theta_\text{thresh}$ (Fig.~\ref{fig:tag_descriptor}.3).
For temporal matching and dimensionality reduction, the points in $\mathcal{B}^{'}_{t-1}$ and $\mathcal{B}_{t}$ are assigned to clustering groups $\mathcal{G}^i_{t-1}$ and $\mathcal{G}^i_t$. 
This assignment is based on the Euclidean distance to their clustering center $\mathbf{g}^i$, within a fixed threshold $d_\text{thresh} = 0.25 \si{\meter}$ (Fig.~\ref{fig:tag_descriptor}.4).
Note that $d_\text{thresh}$ is a static parameter and chosen with a safety margin based on the relation between maximum expected obstacle speed and the inference time step as $d_\text{thresh} > v_\text{max} \Delta t$.
The 2D~centroids \mbox{$\mathbf{c}^i$} of each group $\mathcal{G}^i_{t-1}$ and $\mathcal{G}^i_t$ counteract sensor noise and finally represent a single TAGD \mbox{$(\mathbf{c}_t^i, \mathbf{c}_{t-1}^i)$} (Fig.~\ref{fig:tag_descriptor}.5).

A TAGD represents the center of data points across two consecutive lidar scans close to the nearest obstacle within an angular zone, such that even small obstacles hit by only a single ray are successfully represented by a TAGD.
Note that it is possible for a single dynamic obstacle to be represented in more than one TAGD, depending on the positions of clustering centroids, e.g., see TAGDs 26 and 27 in Fig.~\ref{fig:tag_descriptor}.4).
However, such double representations did not hinder the performance in context of the learned controller.
With regards to real-world pedestrians and their leg motion pattern, the influence of faster-than-body moving single legs on the TAGD displacement and therefore body speed estimation cannot be entirely ruled out.
However, the group centroid calculation within $d_\text{thresh}$ supports averaging out the effects of sensor noise or displacement and speed of individual legs, even though leg walking patterns are not explicitly considered or simulated in this work.
It is worth noticing that a consistent inference timing between the lidar scans is key to correctly represent a given obstacle velocity with TAGDs, also with regards to a sim-to-real transfer.
Here, this is directly based on the reinforcement learning control time step $\Delta t = \num{0.2}\si{\second}$.
We have not used odometry or IMU data for enhanced ICP alignment, but solely rely on the observed static obstacles in the scene.
While posing a limitation, this is a defensible assumption for indoor environments.
However, we will evaluate the reliance and performance dependency of the navigating RL agent on correct ICP alignment in two ways, 1) without static obstacles in open space among dynamic obstacles, and 2) with ICP alignment artificially turned off.
In summary, TAGDs reveal obstacle motion and will therefore be used as input to the temporal attention module of our pipeline.
\vspace*{-0.75em}
\begin{algorithm}[!hb]
	\caption{Temporal Accumulation Group Descriptors \label{algo:tagd}}
	\begin{algorithmic}
		\Require Lidar readings $\mathcal{B}_{t-1}, \mathcal{B}_{t}$, $d_\text{thresh}, d_\text{max}, N_c$
		\State $\theta_\text{thresh} \gets \pi/N_c$	
		\State $\mathcal{B}^{'}_{t-1} \gets \text{ICP}(\mathcal{B}_{t-1}, \mathcal{B}_{t})$ 
		\State Initialize TAGD list $\mathcal{C}_{t} = \{\}$
		\For{$i=0$ to $N_c$}
			\State $\theta_\text{ref} \gets 2\pi i / N_c$ 
			\State $T \gets \{\mathbf{b}=(r, \theta) \in  \mathcal{B}_{t}| \text{ }|\theta - \theta_\text{ref}| \leq \theta_\text{thresh}\}$
			\State $r_\text{min} \gets \min_{(r, \theta) \in T}(r, d_\text{max})$
			\State $\mathbf{g}_i \gets (r_\text{min}, \theta_\text{ref})$
			\State $\mathcal{G}^i_t \gets \{ \mathbf{b} \in \mathcal{B}_{t} \,|\, \text{dist}(\mathbf{b}, \mathbf{g}_i) \leq d_\text{thresh} \}$ 
			\State $\mathcal{G}^i_{t-1} \gets \{ \mathbf{b}^{'} \in \mathcal{B}^{'}_{t-1} \,|\, \text{dist}( \mathbf{b}^{'}, \mathbf{g}_i) \leq d_\text{thresh} \}$
			\State TAGD $(\mathbf{c}_t^i, \mathbf{c}_{t-1}^i) \gets (\text{centroid}(\mathcal{G}^i_t), \text{centroid}(\mathcal{G}^i_{t-1}))$ 
			\State $\mathcal{C}_{t} \gets \mathcal{C}_{t} \cup \{(\mathbf{c}_t^i, \mathbf{c}_{t-1}^i)\}$
		\EndFor			
	\end{algorithmic}
\end{algorithm}

\vspace*{-1.6em}
\subsection{Deep Reinforcement Learning for Navigation}
\vspace*{-0.25em}
We choose a deep deterministic policy gradient (DDPG) architecture consisting of an actor and a critic, modeled by neural networks \cite{lillicrap2015continuous}. 
DDPG features a continuous action space, allowing for smooth robot control. 
The actor network outputs linear and angular velocities for the robot.
The RL framework is based on the Markov Decision Process:
An agent in state $s_t$ at time step $t$ decides upon an action $a_t$ based on a policy $\pi(s_t) = a_t$. 
Upon reaching the next state $s_{t+1}$, it receives a reward $r_t$.
The optimization objective is the maximization of the \mbox{$\gamma$-discounted} cumulative return \mbox{$R = \sum_{i=t}^{T}\gamma^{i-t}r_t$}, where \mbox{$\gamma=0.98$}.
As DDPG is an off-policy RL algorithm, the state-action pairs are stored in a experience replay buffer of length \mbox{$N_{\text{RP}} = 2,000,000$} and sampled in batches for policy updates.

\vspace*{-0.5em}
\subsection{State and Action Space}
\begin{figure*}[!ht]
	\includegraphics[width=1.0\linewidth]{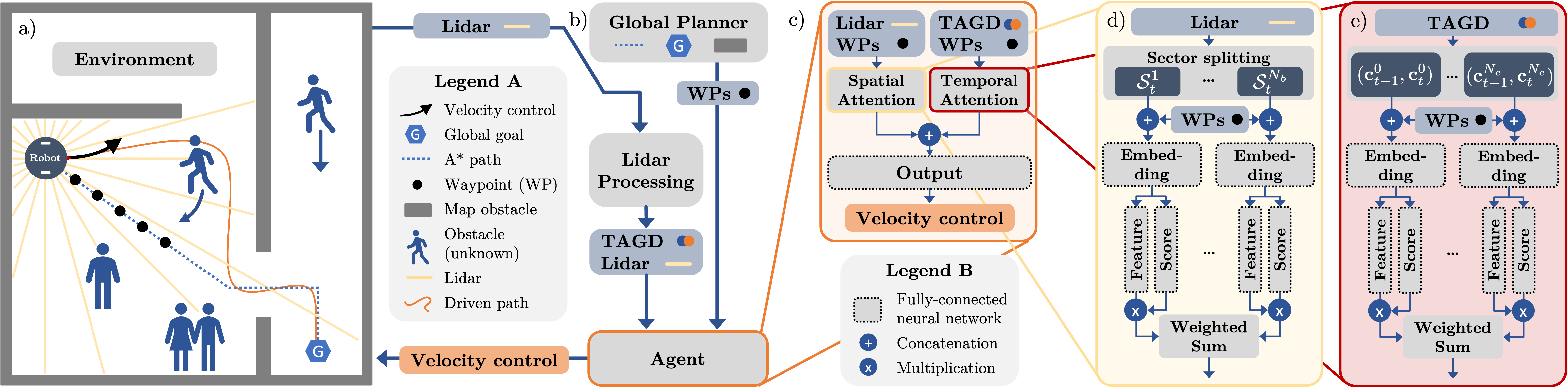}
	\centering
	\caption{
		Illustration of our architecture.
		\textbf{a)} The indoor environment provides lidar readings to the deep reinforcement learning agent that drives a differential-wheeled robot via linear and angular velocity commands.
		\textbf{b)} From subsequent lidar readings, the TAGDs are computed. 
		Merged with the five upcoming waypoints of the global path and the raw lidar readings as observations, they are 
		\textbf{c)} processed by the agent in a separate spatial and temporal stream. 
		Both streams feature an attention block to weigh the importance of \textbf{d)} individual lidar sectors (spatial) or \textbf{e)} the TAGDs (temporal), with respect to the upcoming waypoints.
		After feature extraction, both streams are concatenated for further processing in the final output network of the actor-critic agent.
	}
	\label{fig:network_architecture}
\end{figure*}

\vspace*{-0.25em}
The state space defines the observations we provide to the agent. 
As can be seen in Fig.~\ref{fig:network_architecture}b, the agent has access to 2D~lidar sensor data, the lidar-derived TAGDs, and upcoming waypoints for global guidance:
The $N = 180$ ray min-pooled lidar scan $\mathcal{B}_{t}$ is represented as a set of robot-centric Cartesian 2D~points.
Focusing on local obstacles, the lidar scanning range is limited to $d_\text{max} = \num{3.5} \si{\meter}$.
To sense dynamic obstacles, we provide the TAGDs \mbox{$\mathcal{C}_{t} = \{(\mathbf{c}_t^i,  \mathbf{c}_{t-1}^i) | 0 \leq i < N_c\}$} as described earlier, where the number of TAGD's $N_c=30$ is equal to the number of spatial sectors $N_b=30$.
The value $N_c=30$ was heuristically chosen so that the clustering groups would jointly provide a cohesive circular coverage
mid-range of the lidar distance around $d=2.5\si{\meter}$, and the TAGD clustering circles defined by the radius $d_\text{thresh}$ start to overlap $(N_c \approx 2\pi d/(2 d_\text{thresh}))$.
From the robot nearest waypoint $\mathbf{p}^{c}$ on path $\mathcal{P}$, we sample $N_f = 5$ waypoints spaced at  \mbox{$\Delta \mathbf{p}^i = \num{0.3} \si{\meter}$} towards the goal. These are converted to robot-centric Cartesian coordinates and input to the agent as \mbox{$\mathcal{P}_t^f = \{\mathbf{p}^i | c \leq i < c + N_f \}$}.
		
The continuous action space of the agent consists of linear and angular velocities $(v, w)$, with a range of \mbox{$v \in [0, 1.0] \si{\meter\per\second}$} and \mbox{$w \in [-\pi, \pi] \si{\radian\per\second}$}.
The robot is not allowed to drive backwards to foster foresighted navigation.

\vspace*{-0.5em}
\subsection{Reward}
\vspace*{-0.25em}
\label{sec:reward}
The navigating agent's overall objective is to navigate collision-free along a given path among unknown dynamic obstacles.
The reward $r_t$ is therefore a weighted sum:
\begin{equation}
	\label{eq:reward_path} 
	r_t = \alpha_1 r^{\text{collision}}_t + \alpha_2 r^{\text{guide}}_t  + \alpha_3 r^{\text{prox
	}}_t
\end{equation}
$\alpha_1= 10$, $\alpha_2= 0.2$ and $\alpha_3 = 3$ are the experimentally determined weighting factors.

To encourage collision-free navigation, we penalize with $r^{\text{collision}} = -1$ upon collision of the robot with any obstacles.

A natural guidance along the global path is beneficial as it encourages the agent to drive towards the goal. 
From the current closest waypoint $\mathbf{p}^{c}$ on the path to the robot, we interpolate $0.6\si{\meter}$ forward along the path to obtain the guidance point $\mathbf{p}^g$.
The distance between $\mathbf{p}^g$ and the robot's position $\mathbf{p}^r$ are penalized with \mbox{$r^{\text{guide}} = - ||\mathbf{p}^g - \mathbf{p}^r||$}. 
By design and due to the update at every time step, $\mathbf{p}^g$ cannot be reached, thus providing a continuous penalty that increases when the robot deviates from the path and encourages the robot to drive back to the path in a forward-leading manner.

The concept of $r^{\text{prox}}$ aligns with the sparse collision reward, but does not terminate the episode for easier learning.
Instead it alerts the agent in vicinity to obstacles about higher risk of collisions, or in other words encourages the agent to keep clear of obstacles.
When the minimum distance $d$ between robot and any lidar-scanned obstacle falls below $d_{\text{prox}}= \num{0.5} \si{\meter}$, a linearly growing penalty is computed as  \mbox{$r^{\text{prox}} =  -1 \times \vert d_{\text{prox}} - \min (d, d_{\text{prox}}) \vert$}, else \mbox{$r^{\text{prox}}=0$}.

\vspace*{-0.75em}
\subsection{Network Architecture}
\vspace*{-0.25em}
\label{sec:network_architecture}
As shown in \figref{fig:network_architecture}c, our agent's architecture is constructed around two data streams.  
The individual streams extract spatial and temporal features via an attention mechanism, respectively.
Note that both the down-sampled lidar input of the spatial, and the TAGD input of the temporal steam contain partially redundant information due to their origin in the raw distance readings.

\subsubsection{Temporal and Spatial Data Stream}
With the individual TAGDs and the possibly attention-relevant, therefore redundantly-represented upcoming waypoints, we construct $N_c$ individual vectors $\mathbf{U}_{temp} = \{[\mathbf{c}_{t-1}^i, \mathbf{c}_{t}^i, \mathcal{P}^f_t] | 0 \leq i < N_c\}$, to be passed on to the temporal attention module as $\mathbf{y}_{temp} = Att_{temp}(\mathbf{U}_{temp})$, see Fig.~\ref{fig:network_architecture}e.
In the spatial data stream, the lidar scan $\mathcal{B}_t$ of \mbox{$N$} rays at time step $t$ is split into \mbox{$N_b$} angular sectors $\mathcal{S}_t^i = \{\mathbf{b}_t^j \in \mathcal{B}_t | iN_b\leq j < (i+1)N_b\}$ with \mbox{$N/N_b$} rays each.
Again, each sector-vector  is concatenated with next path segment forming $N_b$ individual vectors $\mathbf{U}_{spat} = \{[\mathcal{S}_t^i, \mathcal{P}^f_t] | 0 \leq i < N_b\}$, jointly passed on to the spatial attention module as $\mathbf{y}_{spat} = Att_{spat}(\mathbf{U}_{spat})$.
After both data streams have been processed by their attention modules, respectively, they are concatenated and jointly processed by an output module $O(\cdot)$ for joint feature extraction $\mathbf{o} = O(\mathbf{y}_{temp}, \mathbf{y}_{spat})$.
Two separate modules of this pipeline form the actor and critic.
\subsubsection{Attention module}
\label{sec:attention_module}
Both temporal and spatial attention modules $Att(\cdot)$ share a similar network architecture, but no parameters.
A visualization of our lightweight attention module can be found in \figref{fig:network_architecture}d-e. 
It is constructed with an embedding, a score and a feature network, inspired by Chen\etal \cite{chen2019crowd} and \cite{de_heuvel_subgoal-driven_2023}.
The embedding module $E(\cdot)$ encodes the input vectors individually along the attention dimension to $\mathbf{e}_i = E(\mathbf{u}_i)$. 
The embedding $\mathbf{e}_i$ is fed into the score module $S(\cdot)$ that outputs the attention scores $\mathbf{s}_i = S(\mathbf{e}_i)$. 
All attention scores are Softmax-normalized to obtain the final importance weight.
In parallel the embedding is also fed into the feature module $F(\cdot)$ that generates the feature representations as $\mathbf{f}_i = F(\mathbf{e}_i)$. 
Finally, the feature vectors are scaled by their importance in a weighted sum.
\vspace*{-0.25em}
\begin{align}
	\mathbf{y} = Att(\mathbf{U})	&= \textstyle \sum_i \text{Softmax}(\mathbf{s}_i) \cdot \mathbf{f}_i\\
						&= \textstyle \sum_i \text{Softmax}(S(\mathbf{e}_i)) \cdot F(\mathbf{e}_i)
\end{align}
\vspace*{-0.25em}
Note that the due to the lightweight implementation of our attention scheme, the dimensionality along the attention axis reduces from $N_b$ or $N_c$ vectors to one in the output.
In other words, the individually embedded lidar sectors or TAGDs do not attend to each other, but the attention scales their impact in the weighted sum, respectively.
This form of attention is also referred to as \textit{location-based} attention \cite{niu_review_2021, luong_effective_2015-1}.
All networks describe above are constructed with as ReLU-activated multi-layer perceptrons (MLP)\footnote{Layer sizes (hidden nodes): embedding: $256 \times 128 \times 64$, score: $60 \times 50 \times 1$, feature: $80 \times 50 \times 30$, output: $128 \times 64 \times 64 \times \{1,2\}$ (critic/actor)}. 
\vspace*{-0.5em}
\subsection{Indoor Training Environments}
\vspace*{-0.25em}
\label{sec:environments}
\begin{figure}[t]
	\centering
	\includegraphics[width=0.9\linewidth]{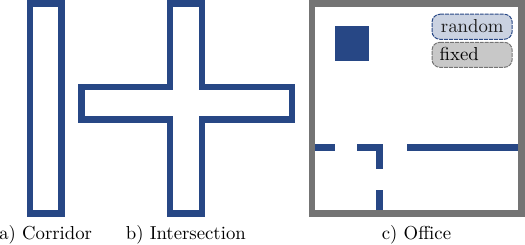}
	\caption{
		The Pybullet-based environments of \cite{de_heuvel_subgoal-driven_2023} are used for training.
		\textbf{a)} In the corridor and  \textbf{b)} intersection environment, the wall distances are randomized (blue). 
		\textbf{c)} In the office environment, the outer walls are fixed with randomized inner wall placement for diverse room setups.
	}
	\label{fig:environments}
\end{figure}
To train our navigation agent, we use the Pybullet\cite{coumans2019} physics engine.
We use the minimalistic but well-randomizing indoor environments of de Heuvel\etal \cite{de_heuvel_subgoal-driven_2023} featuring dynamic cuboid obstacles that represent pedestrians, with three different types of scenarios, see Fig.~\ref{fig:environments}: 
Corridors, intersections, and offices.
The randomization of wall density and placement provides varying levels of scene complexity. 
The corridor environment is long and narrow with a length between $[ 6\si{\meter}, 8 \si{\meter}]$ and a width between $[ 2.0\si{\meter}, 2.5\si{\meter} ] $. The robot encounters pedestrians moving in opposite directions.
The intersection environment is cross-shaped featuring hallway widths between $2.0\si{\meter}$ and $2.5\si{\meter}$, and includes corners that create blind spots for sudden pedestrian appearances.
The office environment features a fixed outer size with randomized interconnected rooms and introduces doorway encounters where the robot waits for pedestrian clearance before proceeding.
Our room types cover typical encounters suggested for social navigation tasks \cite{francis_principles_2023}, as found also in other related studies \cite{perez-darpino_robot_2021, liang_realtime_2020}.
While our rectangular environments generate variety through architectural randomization, other works achieve variety through larger but static, non-rectilinear scenes \cite{fan2019getting}.
The robot's start and goal location are sampled  in the corners or dead ends of the scenes, respectively.

\subsubsection{Obstacle simulation}
Dynamic and static pedestrians represented by cuboids move back and forth through the environments along A* paths with randomized quantity ($N_\text{dyn} \in [1, N_\text{dyn}^\text{max}]$, $N_\text{stat} \in [1, 2]$), speed ($v_{\text{ped}} \in [0.5, 1.0] \si{\meter\per\second}$), start, and goal position.
Note that the pedestrian speed can exceed the robot's maximum velocity.
The maximum dynamic obstacle number $N_\text{dyn}^\text{max} \in \{2, 4, 8\}$  follows a curriculum scheme (three levels) and is increased over the course of training, whenever the evaluation success rate exceeds $70 \si{\percent}$.
For the purpose of increasing the obstacle encounter likelihood with the robot, start and goal locations of the first pedestrians are sampled around the robot path.
All other pedestrians will cross the robot path eventually.
Note that the A*-following pedestrians do not take into account each other or the robot position, but rigorously move forward.
Collision avoidance is therefore entirely up to the robot, similar to \cite{fan2019getting, liang_realtime_2020}
This can lead to highly challenging navigation encounters, especially for larger obstacle numbers.
This is in contrast to other studies \cite{perez-darpino_robot_2021} that simulate the pedestrians motion based on Optimal Reciprocal Collision \mbox{Avoidance (ORCA)}, where the pedestrians avoid each other. 
Notably, also the robot is actively avoided by the pedestrians, easing the collision-free navigation task for the RL agent.
Other works have employed the social force model for crowd navigation \cite{kolivand_integration_2021}.
Though our more basal dynamic obstacle simulation leads to occasional pedestrian mesh overlaps and occasionally non-passable situations, our selection of an only path-based model is justified by our study's primary focus on feature extraction for RL-driven dynamic obstacle avoidance, rather than on crowd navigation.
	
\vspace*{-0.6em}
\subsection{Robot Model}
\vspace*{-0.25em}
\label{sec:robot}
We employ a differential-wheeled robot, more precisely, the Kobuki Turtlebot 2.
The Turtlebot performs angular turns with a speed difference between both wheels.
A Slamtec RPlidar A3 2D~lidar sensor is mounted on top of the Turtlebot, emitting 1,440 beams.
In simulation, we add sensor noise to the distance readings with an amplitude of $\num{2.5} \si{\centi\meter}$.

\vspace*{-0.4em}
\section{Experiments}
\vspace*{-0.25em}
\label{sec:exp}
In the following we present the training and evaluation details, followed by an ablation and baseline study.
After evaluating the domain shift to the iGibson simulator, the section is rounded up by the real-robot deployment.

\vspace*{-0.6em}
\subsection{Training Setup}
\vspace*{-0.25em}
An episode denotes one navigation run of the robot from start until one of the termination criteria is reached: 
Collision with other obstacles, timeout after $T_\text{timeout} = 150 \equiv 30 \si{\second}$ steps, or goal-reaching upon vicinity of $0.2 \si{\meter}$ to the global goal.
To foster generalization abilities, for each episode a randomly generated environment is setup, as described in Sec.\ref{sec:environments}.
The inference and control time step of the agent is set to $\Delta t = \num{0.2} \si{\second}$, which also represents the time difference between subsequent lidar scans for the temporal processing.
The learning rates for both actor and critic is $\num{1e-4}$.
All agents presented are trained for \mbox{300,000} episodes and evaluated regularly.
The best performing model checkpoint of the highest curriculum level is selected for all approaches.

\vspace*{-0.75em}
\subsection{Quantitative Performance}
\vspace*{-0.25em}
\begin{figure}[!t]
	\centering
	\includegraphics[width=0.95\linewidth]{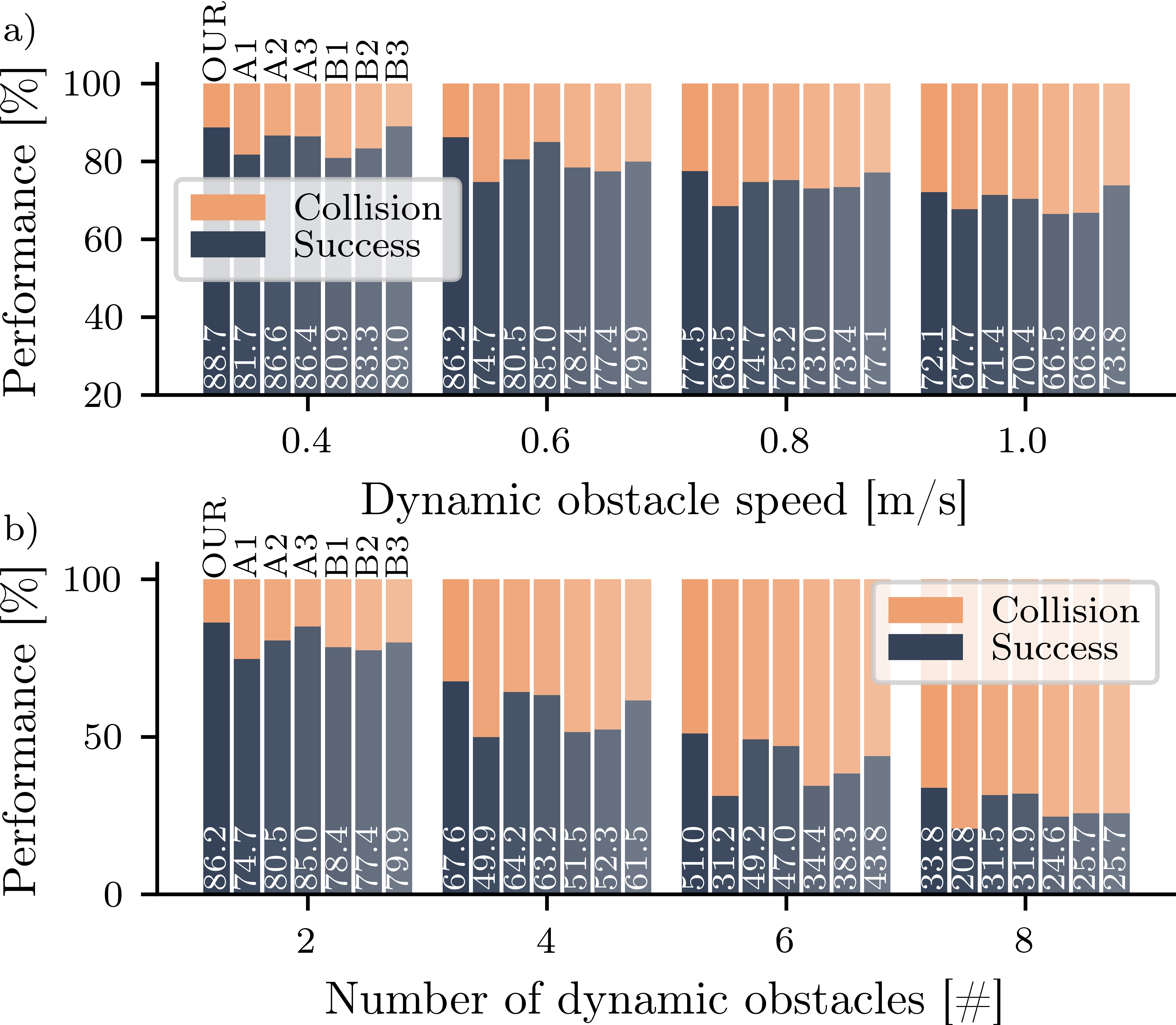}
	\caption{
		Performance overview for all approaches averaged over \mbox{1,000 episodes} with identical scene setups in all three Pybullet environments for 
		\textbf{a)} increasing obstacle speeds, with two dynamic and one static pedestrians, and 
		\textbf{b)} increasing number of obstacles, with a fixed pedestrian speed $0.6 \si{\meter\per\second}$.
		} 
	\label{fig:results_compare}
\end{figure}

We evaluated our trained models with respect to success rate, collision rate, timeout rate, and navigation time over \mbox{1,000} episodes.
For comparability, the \mbox{1,000} episodes were setup identically among all approaches.
The flagship approach presented in this study is denoted with OUR.
Generally, with challenging environment complexity due to increased obstacle velocities (Fig.~\ref{fig:results_compare}a), or increased number of dynamic obstacles (Fig.~\ref{fig:results_compare}b), the success rate stagnates. 

\begin{figure}[!t]
	\centering
	\includegraphics[width=1.0\linewidth]{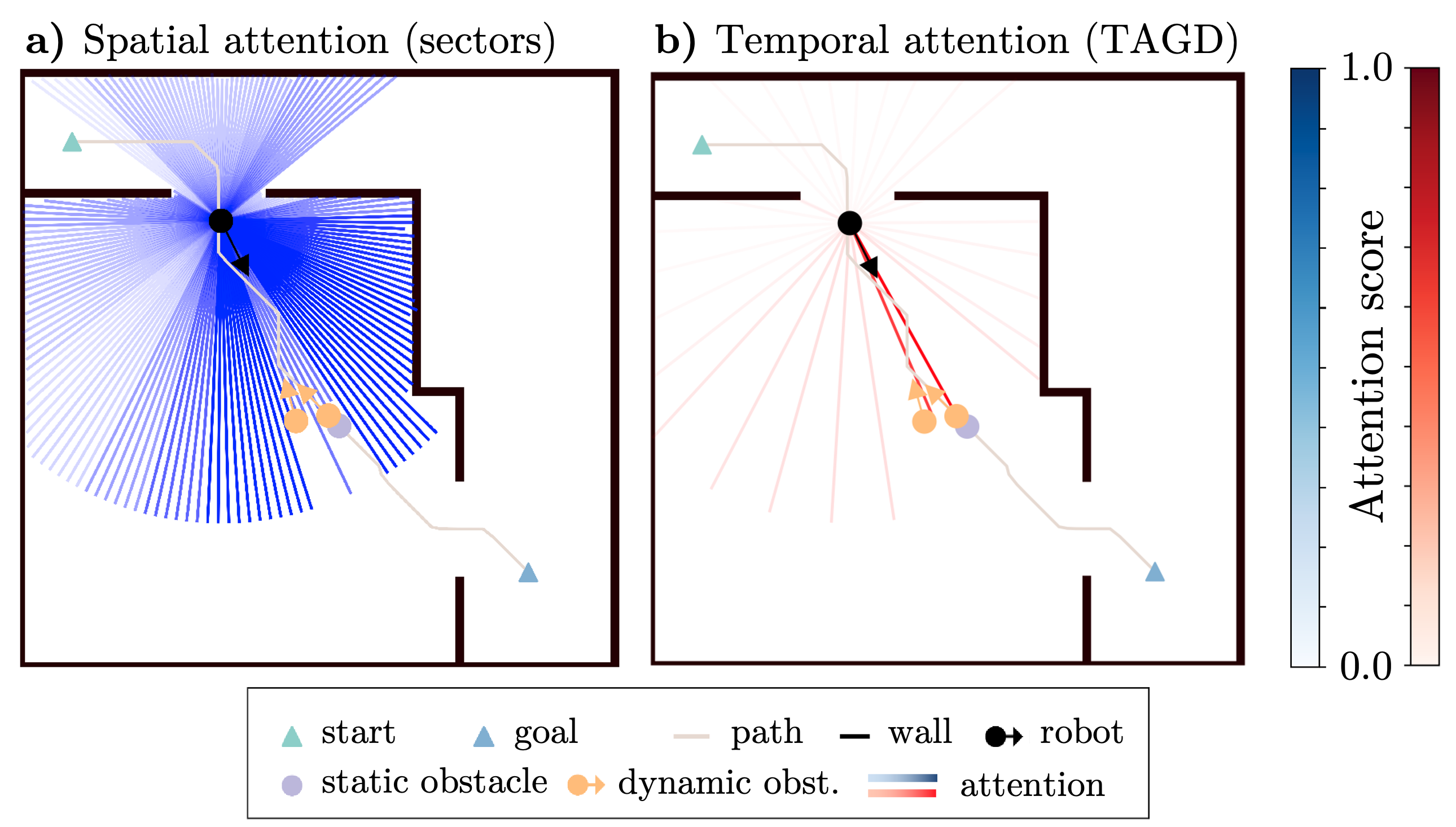}
	\caption{
		Exemplary visualization of the \textbf{a)} spatial (blue) and \textbf{b)} temporal attention (red) for a given navigation scene. 
		The attention scores were color-mapped onto the lidar beam sectors for the spatial and on the beams pointing towards the TAGDs for the temporal attention, respectively.
		Increased spatial attention towards the forward-facing lidar sectors, as well as increased temporal attention towards the oncoming dynamic obstacle can be observed.
		}
	\label{fig:scene_and_attention}
\end{figure}
\vspace*{-0.15em}
\subsubsection{Ablation Study}
We did an ablation study with respect to OUR approach described above to evaluate the contribution of each module to the results, see Tab.~\ref{tab:ablation}a) and Fig.~\ref{fig:results_compare}.

A1 NO-SPATIAL: As OUR, but removing the spatial attention stream, leaving only TAGD and waypoint processing.

A2 NO-TEMPORAL: As OUR, but with no temporal stream or TAGD input, leaving only the spatial single time step attention stream and waypoint processing.

A3 NO-TAGD: As OUR, but without TAGD preprocessing. 
The network structure implements the spatial attention stream twice with separate network parameters, each processing one of the consecutive lidar scans, respectively.

As can be seen from Tab.~\ref{tab:ablation}a) and Fig.~\ref{fig:results_compare}, with all ablations the performance deteriorates.
The joint contribution of spatial and temporal attention emerges with A2 NO-TEMPORAL having a lower success rate compared to OUR, as it relies only on single-time step spatial information.

\setlength\lightrulewidth{0.1ex}
\begin{table}[b!]
	\centering
	\begin{tabular}{llcccc}
		\textbf{a)} & \textbf{Ablation} & SR$\uparrow$ & CR$\downarrow$ & TR$\downarrow$ & Nav. time$\downarrow$ \\
		\toprule
		&OUR 						& \textbf{86.2}  	& \textbf{13.8} 	& 0.0  	& \textbf{17.7} $\si{\second}$\\
		&A1: NO-SPATIAL 				& 74.7  			& 25.3 				& 0.0  	& 18.1 $\si{\second}$\\
		&A2: NO-TEMPORAL 			& 80.5  			& 19.5  			& 0.0  	& 17.9 $\si{\second}$ \\
		&A3: NO-TAGD 				& 85.0  			& 15.0  			& 0.0  	& 17.8 $\si{\second}$\\
		%
		\textbf{b)} & \textbf{Baseline} & & & & \\
		\midrule[\lightrulewidth]
		&B1: Liang\etal \cite{liang_realtime_2020} 				& 78.4  & 21.6  & 0.0  & 18.9 $\si{\second}$\\
		&B2: Pérez-D.\etal \cite{perez-darpino_robot_2021} 		& 77.4  & 22.6  & 0.0  & 18.6 $\si{\second}$\\
		&B3: Pérez-D.\etal \cite{perez-darpino_robot_2021} 		& 79.9  & 20.1  & 0.0  & 18.4 $\si{\second}$\\
		%
		\textbf{c)} & \textbf{Generalization} & & & & \\
		\midrule[\lightrulewidth]
		&iGibson \cite{shen_igibson_2021-1} & 79.2 & 18.6 & 2.2 & 19.0 $\si{\second}$\\
		\bottomrule
	\end{tabular}
	\caption{
		Performance rates in [$\si{\percent}$] with respect to success (SR), collision (CR), and timeout (TR) and average navigation times for successful episodes of \textbf{a)} ablation and \textbf{b)} baseline study averaged over 1,000 episodes, with $2$ dynamic pedestrians ($0.6 \si{\meter\per\second}$) and $1$ static pedestrian. 
		The \textbf{c)} generalization evaluation reveals slightly decreased performance for the post-training domain shift to the iGibson simulator on similar navigation tasks in more complex environments.
		}
	\label{tab:ablation}
\end{table}

\vspace*{-0.6em}
\subsection{Baselines}
\vspace*{-0.25em}
To identify the contribution of our feature extraction approach, we compared against two baseline architectures.
All baselines leverage 2D~lidar ($\num{360} \si{\degree}$) for learning-based mobile robot navigation and were trained in the same environment and training parameters as our approach.
The baseline-related modifications lie in the state space content and processing network architectures.

\subsubsection{Liang\etal - B1}
A highly-related state-of-the-art approach has been presented in \cite{liang_realtime_2020}. 
Similarly to ours, it is an end-to-end obstacle avoidance algorithm originally trained with Proximal Policy Optimization.
The authors use \mbox{2D lidar} and a depth camera to perceive the environment, while the controller outputs velocity commands. 
From both perception modalities, we solely implement the lidar-related preprocessing and network architecture to replace our attention blocks, which is a \mbox{1D CNN} taking in three consecutive scans.
Precisely, this module is composed of two \mbox{1D CNN} layers followed by a fully-connected MLP.
In contrast to our approach with 2D~Cartesian point lidar representation, single-value lidar distance readings are used.
The state space still contains five upcoming waypoints, which in contrast to OUR are processed by a separate MLP.
Without convergence and therefore not included, we have also tested a closer-to-the-original implementation (512 lidar rays, no waypoints, only goal position).

\subsubsection{Pérez-D'Arpino\etal - B2/3}
In the end-to-end lidar navigation approach of \cite{perez-darpino_robot_2021}, no temporal information but only the current lidar reading is processed.
Similar to Liang\etal \cite{liang_realtime_2020}, the authors employ a lidar-processing 1D CNN but with three layers followed by a fully-connected layer.
Furthermore, $N=\num{128}$ single-value lidar distance readings are used.
Additionally, the global goal position and next upcoming waypoints of the A* path ($\Delta \mathbf{p}^i = \num{1.0} \si{\meter}$) are part of state space.
In B2, we employ their state space and replace our attention block with their lidar-processing CNN and waypoint-processing MLP modules.
A sub-version (B3) uses only their CNN architecture but our original state space with regards to waypoints and lidar resolution.

\subsubsection{de Heuvel\etal}
In initial tests, we compared against \cite{de_heuvel_subgoal-driven_2023}, outputting collision-free subgoals instead of velocity commands from single-time step lidar data with similar spatial attention.
Direct comparisons with our current method are not viable due to later changes in the training settings.
Despite this, the comparison showed a $\num{5,3} \si{\percent}$ performance boost by incorporating TAGDs and temporal attention, motivating our current work.

As can be seen in Tab.~\ref{tab:ablation}b) and Fig.~\ref{fig:results_compare}, for our setup, the CNN-related baselines \mbox{B1-B3} struggle more with increased number of obstacles.
In almost all cases, our approach outperforms all baselines in terms of success rate.

\begin{figure*}[ht]
	\includegraphics[width=1.0\linewidth]{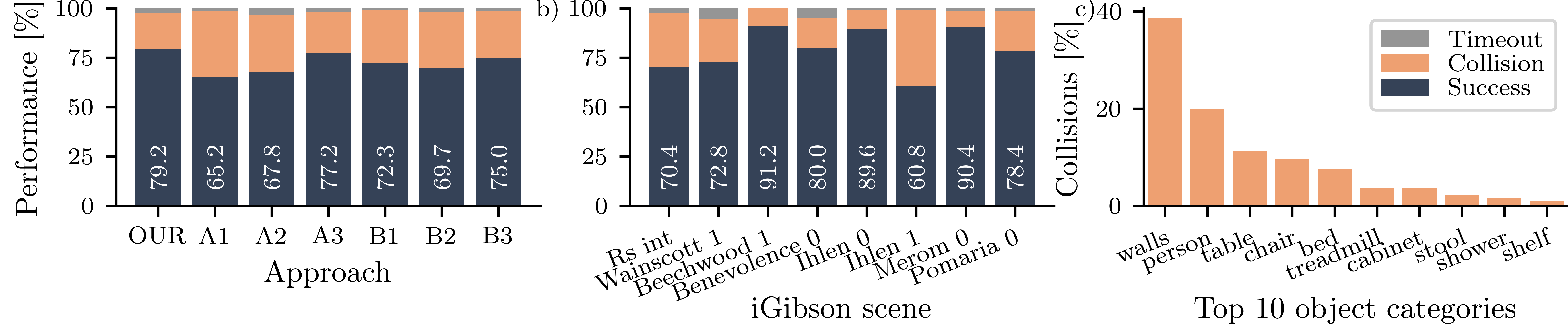}
	\centering
	\caption{
		Results of the generalization study using the iGibson simulator over eight scenes witch $125$ episodes each.
			\textbf{a)} OUR controller demonstrates the best generalization capabilities in the sim-to-sim transfer of all approaches.
			\textbf{b)} Breakdown into the different scenes shows a scene-dependency of the controller performance for OUR controller.
			\textbf{c)} Collision object category analysis within iGibson: The most collided-with objects are the walls.
	}
	\label{fig:igibson}
\end{figure*}
\vspace*{-0.6em}
\subsection{Qualitative Attention Analysis}
\vspace*{-0.25em}
\figref{fig:scene_and_attention} visualizes the learned spatial (a) and temporal (b) attention for a given navigation scenario. 
Here, two dynamic obstacles approach the robot from opposite directions, the robot has just entered the room.
The spatial attention highlights the forward lidar sectors in the desired direction of navigation.
The robot navigates along a wall that locates on its left hand side and we can observe an increased attention on the corresponding lidar sectors.
Intuitive to the human eye, the temporal attention focuses the TAGDs of the oncoming dynamic obstacles.
Similar to the spatial attention, a slightly increased temporal attention can be observed in forward direction of the robot.
In direct comparison to the temporal stream, the spatial stream exhibits a less sharp attention distribution in this scenario.
Further attention visualizations can be found in the accompanying video\footnote{\href{https://youtu.be/cYNUFD_rGNE}{https://youtu.be/cYNUFD\_rGNE}\label{fn:video}}.

\vspace*{-0.6em}
\subsection{Robustness}
\vspace*{-0.25em}
Verifying the robustness of our approach with respect to ICP accuracy against its dependence on static obstacles for correct alignment, an open space evaluation of the same evaluation environments but without walls reveals an absolute performance decrease of $3.6 \si{\percent}$. 
When disabling ICP alignment entirely and feeding non-aligned lidar scans into the TAGD pipeline,  the absolute performance drops by $4.8 \si{\percent}$.
In both cases, the performance is still superior the NO-TEMPORAL ablation, demonstrating decent robustness of the TAGD-based approach against ICP failure in these edge cases. 
Note that the obstacle parameters of Tab.~\ref{tab:ablation} were used.

\vspace*{-1.3em}
\subsection{Generalization Performance}
\vspace*{-0.25em}
To investigate the generalization ability of our approach, we evaluated the Pybullet-trained agents in the iGibson simulator \cite{shen_igibson_2021-1} in a sim-to-sim transfer, see Fig.~\ref{fig:motivation}.
The sensor settings and overall navigation objective remain similar, but two major differences strike:
1) The indoor scenarios are of high fidelity with diverse furniture objects and a more complex room architecture.
2) The pedestrians are represented with real \mbox{3D meshes} instead of cuboids and have a more refined motion simulation.
Precisely, we adapt the navigation task from the 2021 iGibson Social Navigation Challenge \cite{noauthor_igibson_nodate} that features eight scenes and Optimal Reciprocal Collision \mbox{Avoidance (ORCA)} among pedestrians.
The key settings to mention as taken over from the original challenge are the maximum pedestrian speed of $0.5 \si{\meter\per\second}$, an inverse scene area-related population of 8 $\si{\square\meter}$ per pedestrian, and a goal sample distance between 1.0 and $10.0 \si{\meter}$.

As seen in Fig.~\ref{fig:igibson}a), OUR controller exhibits the best generalization performance among all approaches.
The slightly lower success rates in Fig.~\ref{fig:igibson}a) and Tab.~\ref{tab:ablation}c) point towards a simulator gap and increased difficulty within the scenes.
Also, the individual scenes seem to be of varying difficulty to the robot, compare Fig.~\ref{fig:igibson}b).
To further differentiate the challenges the robot faces in the iGibson scenes, the top ten collided-with object categories have been recorded, see Fig.~\ref{fig:igibson}c).
As the majority of collisions events involve walls, the possibly higher degree of confined spaces within the iGibson scene could play a role.
Furthermore, tables and chairs are among the most frequent collision causes.
These object are usually thin-legged, providing a challenge for lidar detection at low angular resolutions.
In summary, the attention-based architecture surpasses the tested CNN feature extractors in unseen environments.
  
 \vspace*{-0.5em}
\subsection{Real-World Experiment}
\vspace*{-0.25em}
Using the Robot Operating System (ROS)\cite{quigley_ros_2009}, we transferred the~trained controller to a real Kobuki Turtlebot 2, as described in Sec.~\ref{sec:robot}.
In our experiment, the Gmapping package \cite{gmapping}, a Simultaneous Localization and Mapping algorithm, was used to build an occupancy grid map of real scenarios upfront for path planning.
During navigation, Adaptive Monte Carlo Localization \cite{fox_monte_1999} estimated the robot's pose in the pre-mapped environment based on the lidar reading and robot odometry. 

We tested our learning-based spatiotemporal approach qualitatively in various real-world scenarios, including corridors, intersections, and offices.
Please refer to our supplemental video\footref{fn:video} of the real-world experiment.
In a corridor, the two participants overtake the robot from behind or approach it rigorously from the front, see Fig.~\ref{fig:motivation}.
The robot smoothly gives room to the pedestrians and avoids collision.
At an intersection, pedestrians appear from the blind spots behind a corner. 
In another test the pedestrian blocks the doorway to see whether the robot would stop upon facing the impassable situation. 
All navigation situations are successfully handled by our spatiotemporal controller.

\vspace*{-0.4em}
\section{Conclusions}
\vspace*{-0.4em}
\label{sec:concl}
We proposed a novel and lightweight approach for robot navigation in dynamic indoor environments. 
Our learning-based approach featuring spatiotemporal attention demonstrates the capacity to highlight collision-relevant features from the sensor data, making the most out of the sparse 2D~lidar readings.
Meanwhile, the introduced temporal accumulation group descriptors (TAGD) help to counteract the robot self-movement over subsequent lidar readings and therefore support the differentiation between static and dynamic obstacles without explicit object tracking.
Our policy directly outputs linear and angular velocity, leading to smooth robot navigation, and outperforms several state-of-the-art approaches in terms of collision rate for different pedestrian speed and number of obstacles. 
We validate the sim-to-sim generalization capabilities in the iGibson simulator, finding excellent and better than state-of-the-art performance to unseen, more complex indoor environments with different pedestrian dynamics.
Lastly, we achieve an effortless sim-to-real transfer into dynamic real-world indoor environments.

\bibliographystyle{IEEEtran}
\bibliography{refs,refs_jorge_resubmission}
\end{document}